\documentclass[10pt, a4paper]{article}

\usepackage{booktabs}
\usepackage[final]{lrec2026} 
\usepackage{tcolorbox}
\usepackage{lineno}

\title{Comparing Natural and  Synthetic Structured Data:\\ A Study of the Passive Verb Alternation in French and Italian}

\name{Giuseppe Samo$^{1}$, Paola Merlo$^{1,2}$} 

\address{$^{1}$IDIAP Research Institute, $^{2}$University of Geneva \\
          \{giuseppe.samo, paola.merlo\}@idiap.ch\\}

\abstract{
This study compares the impact of natural and synthetic data on training and evaluating large language models (LLMs), using the case of passive verb alternation in French and Italian. We use Blackbird Language Matrices (BLMs), structured datasets designed to probe linguistic knowledge of underlying patterns across sentence sets. We compare structured templates instantiated with  natural sentences extracted from Universal Dependencies to structured templates of  synthetic sentences.  Experiments show that while models achieve ceiling performance when trained and tested on synthetic datasets, they do not reliably generalize to natural sentences. In contrast, models trained on natural data exhibit robust performance across both natural and synthetic test suites, demonstrating their superior ability to capture abstract linguistic patterns. These results corroborate the value of natural data and of structured set ups in linguistic evaluation for probing LLMs’ syntactic and semantic knowledge. 
 \\ \newline  }

\begin{document}

\maketitleabstract











\section{Introduction}
\label{introduction}
Whether to use natural or synthetic data for evaluating linguistic knowledge has long been a central methodological debate in theoretical linguistics \citep{chomsky1965,gibson_fedorenko_2013} and computational linguistics (cf. \textit{test suites} vs. \textit{test corpora}; \citealt{lehmann-etal-1996-tsnlp}).

Natural data are easy to define: they are naturally occurring instances -- `wild data'  \citep{bresnan-2016-lifetime} -- extracted from corpora and can be annotated with various types of linguistic information. Synthetic data, by contrast, are artificially generated examples designed to target specific linguistic phenomena under controlled conditions. In this sense, synthetic datasets are conceptually akin to experimental materials in psycholinguistics, where carefully controlled stimuli are constructed to isolate particular grammatical factors \citep{schutze2013judgment,sprouse2017design,futrell-etal-2019-neural}.


To ground this contrast, consider a simple verb alternation, where a verb can appear in a transitive or intransitive form (see also, unspecified object alternation, \citealt{Levin93}), such as the Italian verb \textit{cantare} `sing'. We can colour-code the core elements of the argument structure by marking the constituent with the \textcolor{purple}{Agent} role in purple, the \textcolor{teal}{verb} in teal, and the \textcolor{blue}{Theme} in blue. In a synthetic setting, one can construct minimal, grammar-like examples that clearly instantiate the alternation and control the relevant variables. Consider, for example, the contrast between \textit{\textcolor{purple}{La ragazza} \textcolor{teal}{canta} \textcolor{blue}{la canzone}} (\textit{the girl is singing the song}) and  \textit{\textcolor{purple}{La ragazza} \textcolor{teal}{canta}} (\textit{the girl is singing}). The same structural configurations and their comparison, however, can also be retrieved from naturally occurring corpora by querying dependency relations such as \texttt{nsubj} and \texttt{obj} in annotated treebanks of Universal Dependencies (henceforth UD, \citealt{nivre2015towards,de2021universal}). For example, consider the naturally occurring sentences below, which  instantiate the same relevant constructions:  `\textit{\textcolor{purple}{Mina} \textcolor{teal}{ha} anche \textcolor{teal}{cantato} \textcolor{blue}{molte canzoni di Mogol}}' (\textit{Mina [a singer] has sung many songs of [written by] Mogol [an author]} (ISDT, 2Parole$\_$6-47)) and `\textit{inoltre sino all'ultimo non si sapeva se \textcolor{purple}{il soprano ungherese Andrea Rost}, ammalatasi nei giorni scorsi, \textcolor{teal}{avrebbe potuto cantare} nel ruolo della protagonista Violetta, come alla fine è stato}' (\textit{moreover, until the very end it was not known whether the Hungarian soprano Andrea Rost, who had fallen ill in the previous days, would be able to sing the role of the protagonist Violetta, as in the end she did} (ISDT, isst$\_$tanl-1773)).

\begin{figure}
    \centering
    \begin{tcolorbox}[colback=gray!5,colframe=gray!75!black, width=1\linewidth]
    \footnotesize
    \textbf{Human:} \\
    Can you make two sentences that show the fact that the verb \textit{cantare} in Italian can be both transitive and intransitive?
    
    \vspace{0.1em}
    
    \textbf{DeepSeek:} 
\begin{tcolorbox}[colback=white!5,colframe=gray!75!black, width=1\linewidth]
    \footnotesize
    Certainly! Here are two sentences that demonstrate \textit{cantare} used transitively (with a direct object) and intransitively (without a direct object):
    
    \vspace{0.2em}
    
    \textbf{Intransitive use:} Maria canta sotto la doccia ogni mattina. \emph{(Here, \textit{cantare} is used without a direct object—Maria is simply singing.)}
    
    \vspace{0.2em}
    
    \textbf{Transitive use:} Maria canta una bellissima canzone d'amore. \emph{(Here, \textit{cantare} takes a direct object—\textit{una bellissima canzone d'amore}.)}
   \end{tcolorbox}
    \end{tcolorbox}
    \caption{Example of a synthetic piece of data for transitive and intransitive forms of the Italian verb \textit{cantare} `sing' generated by DeepSeek (DeepSeek-AI, accessed Feb 13, 2026).}
    \label{fig:cantare-examples}
\end{figure}

The notion of synthetic data has broadened considerably in recent years. It includes devised novel examples for experimental purposes, and it also refers to automatically generated outputs from language models and other conversational AI systems \citep{schepens-etal-2025-llmcorpora,Civico2025}. Such data may be synthetic in multiple senses: they are generated automatically rather than naturally occurring, and they can be tailored for experimental purposes. For example, Figure \ref{fig:cantare-examples} shows an interaction in English with DeepSeek V.3. The interaction creates a synthetic instantiation for the alternation of the Italian verb \textit{cantare} by prompting the generation of linguistic examples in English.\footnote{For comparison, ChatGPT (GPT-5-mini, OpenAI) generated the following: ``Intransitive use: \textit{Ogni mattina Maria canta sotto la doccia.} (Here, cantare has no direct object—Maria is simply singing.); Transitive use: \textit{Maria canta una canzone napoletana.} (Here, cantare takes a direct object—una canzone napoletana.). Accessed Feb 13, 2026.} Recent work in NLP has increasingly explored the use of synthetic data to complement or even replace natural data in training and evaluation pipelines \citep{GoyalMahmoud2024,josifoski-etal-2023-exploiting,chen2024diversitysyntheticdataimpact}. However, as noted by \citet{jumelet-etal-2025-multiblimp}, although large-scale annotated resources such as UD have become somewhat marginalized in the era of large language models (LLMs), the extensive linguistic knowledge encoded in these datasets continues to provide an essential foundation for rigorous and linguistically informed evaluation of current language models \citep{opitz-etal2025}.

This paper contributes to this debate by systematically comparing natural and synthetic data within a highly-structured diagnostic test suite targeting specific linguistic phenomena. We examine the extent to which each type of data helps language models identify and abstract the core linguistic elements underlying a complex phenomenon such as a verb alternation—one that cannot be adequately captured by considering sentences in isolation.

We construct structured datasets for the Blackbird Language Matrices (BLM) task \citep{merlo2023blackbirdlanguagematricesblm,merlo2023-findings,an-etal-2023-blm,samo-etal-2023-blm,nastase2024exploring1,nastase2024exploring2,jiang-etal-2024-blm,samo2026datasets}. The BLM task, inspired by Raven’s Progressive Matrices \citep{raven1938}, presents models with sentence-based puzzles designed to capture underlying linguistic rules. Specifically, each task instance consists of a context set of sentences that implicitly encodes the information required to complete a given linguistic pattern, and an answer set of minimally contrasting candidates, only one of which correctly replaces the missing element in the pattern defined by the context. This general template is then instantiated using carefully curated datasets. We instantiate a BLM template in two ways: with natural sentences extracted from UD and with synthetic sentences generated with the help of a ConversationalAI, each in two Romance languages, French and Italian.

The core linguistic phenomenon under investigation is the passive alternation \citep[85]{Levin93} because of its general applicability, across lexical items and languages. Unlike other verb alternations, the passive is not strongly tied to a specific lexical class, but most transitive verbs can undergo passivization (e.g. in Italian with the  passive voice -- colour coded in \textcolor{orange}{orange} -- as in \textit{\textcolor{blue}{La canzone} \textcolor{orange}{è cantanta} \textcolor{violet}{dalla ragazza}} `the song is sung by the girl'), making it a structurally defined alternation,  widely attested cross-linguistically \citep{wals-107}.

Our contribution is two-fold: (i) we introduce multilingual, structured datasets that systematically compare natural and synthetic data with the BLM task and (ii) we evaluate how these data types may shape model representations and their capacity to generalize. This study sheds light on the interaction between training data for structured test suites and linguistic generalization, supporting \citet{jumelet-etal-2025-multiblimp}'s intuition that natural data constitute a robust basis for evaluating language models.

\section{Modelling the Passive Alternation with BLM Paradigms}
\label{task}
\label{lp}

Verb alternations have long been a central object of study in theoretical and computational linguistics, as they provide a window into how syntax, semantics, the lexicon, and the functional lexicon interact \citep{Levin93,rappaporthovav2024variable}.  However, many alternations that are commonly investigated are  lexically specific, only restricted verb classes participate, and cross-linguistic correspondences are often unstable \citep{fillmore2003framenet}. This lexical specificity can  favour superficial patterns in pretrained language models, which may capture surface-level lexical associations rather than abstract syntactic generalizations. The passive alternation, instead, is less restricted in the lexical nature of the verb it can apply to. 

Passive constructions preserve the core argument structure while systematically altering the mapping between syntactic positions and discourse roles \citep{hopper1980transitivity}. The core argument structure of the verb is preserved, but its syntactic realization changes. The verb’s \textcolor{blue}{Theme} is promoted to subject position and, in languages with overt agreement, controls subject–verb agreement, while the \textcolor{violet}{Agent} can be omitted from the surface form, yielding structures that are underlyingly diadic (like the transitive sentence in the introduction) but appear monadic on the surface (like the intransitive sentence in the introduction). 

Across languages, passives are typically signaled by changes in word order and functional morphosyntactic markers. In Italian and French, passive constructions are mainly marked by the presence of an auxiliary verb. Italian typically uses \textit{essere} `be' or \textit{venire} `come', while French primarily uses \textit{être} `be' and, in certain constructions, also \textit{faire} `do'. However, these auxiliaries are not specific to passive structures. For example, in both languages \textit{essere} and \textit{être} `be' also introduce copular and predicative constructions; in Italian, \textit{essere} is additionally used to form compound tenses with certain classes of verbs such as unaccusatives \citep{Burzio1986}. In French, \textit{faire} `do' can also be used in causative constructions. Finally, \textit{venire} `come'  -- as well the other auxiliaries -- function as lexical verbs that can occur in isolation.

The passive form is particularly relevant, as it constitutes a highly pervasive phenomenon in Romance, especially in specific text genres \citep{brunatoetal2022/fpsyg.2022.707630}, while at the same time displaying considerable structural complexity \citep{volpato2010acquisition} and giving rise to effects at the pragmatic level \citep[64]{reinhart1981pragmatics}. As a result, passivization is less fine-grained, and also differently generalizable, than other lexical alternations.


As a  key methodological choice in our study of  alternation phenomena, we move beyond isolated sentences as the primary unit of analysis. Single sentences are often insufficient, since alternations are inherently relational; they concern systematic correspondences between forms rather than properties of individual strings (unlike, for instance, agreement or long-distance dependencies).  Minimal pairs are also a well-established tool in linguistic analysis, but they rely on tightly controlled contrasts and inevitably remain tied to specific lexical items.

For this reason, we adopt a paradigm-based perspective, where the primary object of analysis is a structured set of systematically related constructions. We chose to evaluate the knowledge of passive with the BLM task, which has been shown to be one of the most challenging evaluations for testing Italian knowledge in LLMs \citep{nissim2025challengingabilitieslargelanguage}.


\paragraph{The BLM template}

\begin{figure}
    \centering
  \scriptsize
  \begin{tabular}{l c c p{0.20\textwidth}} 
    \hline
     \multicolumn{4}{c}{\textbf{Template}} \\ \hline
    Structure & \textit{R\textsubscript{Arg}} & \textit{R\textsubscript{ST}} & Example \\ \hline
    \multicolumn{4}{c}{\textsc{Context Set}}  \\ \hline
    1 \textcolor{violet}{Ag} \textcolor{teal}{Vact} \textcolor{blue}{Th} \textcolor{red}{?} & 2 & Q &  \textit{\textcolor{teal}{Does} \textcolor{violet}{the customer} \textcolor{teal}{pay} \textcolor{blue}{the bill}\textcolor{red}{?}}	\\
    2 \textcolor{violet}{Ag} \textcolor{teal}{Vact} \textcolor{blue}{Th} & 2 & D & \textit{\textcolor{violet}{The student} \textcolor{teal}{gets} \textcolor{blue}{the prize}}. \\
    3 \textcolor{violet}{Ag} \textcolor{teal}{Vact} \textcolor{red}{?} & 1 & Q & \textit{\textcolor{teal}{Does} \textcolor{violet}{the teacher} \textcolor{teal}{explain}\textcolor{red}{?}} \\
    4 \textcolor{violet}{Ag} \textcolor{teal}{Vact} & 1 & D & \textit{\textcolor{violet}{The car} \textcolor{teal}{moves}.}\\
    5 \textcolor{blue}{Th} \textcolor{orange}{Vpass} \textcolor{violet}{Ag} \textcolor{red}{?} & 2 & Q & \textit{Why \textcolor{orange}{is} \textcolor{blue}{the case} \textcolor{orange}{studied}}  \\ & &  & \textit{\textcolor{violet}{by the lawyer}\textcolor{red}{?}} \\
   6  \textcolor{blue}{Th} \textcolor{orange}{Vpass} \textcolor{violet}{Ag} & 2 & D & \textit{\textcolor{blue}{The key} \textcolor{orange}{is found} \textcolor{violet}{by the boy}}. \\
   7  \textcolor{blue}{Th} \textcolor{orange}{Vpass} \textcolor{red}{?} & 1 & Q & \textit{When \textcolor{orange}{was} \textcolor{blue}{the screen} \textcolor{orange}{touched}\textcolor{red}{?}} \\
   8 ???  \\
    \hline
    \multicolumn{1}{c}{\textsc{Answer Set}}  \\ \hline
    1 \textcolor{blue}{Th} \textcolor{orange}{Vpass} & 1 & D & \textbf{\textit{\textcolor{blue}{The plants} \textcolor{orange}{were watered}}  }\\
  2 \textcolor{blue}{Th} \textcolor{orange}{Vpass} \textcolor{violet}{Ag} & 2 & D & \textit{\textcolor{blue}{The news} \textcolor{orange}{is reported}} \\
   & & & \textcolor{violet}{by the speaker}.\\
  3   \textcolor{violet}{Ag} \textcolor{teal}{Vact}    & 1 & D & \textit{\textcolor{violet}{The writer} \textcolor{teal}{publishes}.} \\
  4  \textcolor{violet}{Ag} \textcolor{teal}{Vact} \textcolor{blue}{Th} & 2 & D &\textit{\textcolor{violet}{The store} \textcolor{teal}{ships} \textcolor{blue}{the order}.} \\
  5  \textcolor{red}{Question} & 1 or 2 & Q & \textit{How \textcolor{orange}{was} \textcolor{blue}{the data} \textcolor{orange}{analyzed}\textcolor{red}{?}} \\
    \hline
  \end{tabular}
  \caption{BLM template structure instantiated with a synthetic example in English, generated with DeepSeek V.3 (section \ref{syntheticdata}). \textcolor{violet}{Ag} = Agent, \textcolor{blue}{Th} = Theme, \textcolor{teal}{Vact} = verb in active voice, \textcolor{orange}{Vpass} = verb in passive voice, \textcolor{red}{red} elements mark interrogative markers. Number of arguments (1, 2) and sentence type (Q = question, D = declarative).} 
  \label{fig:BLM-template}
\end{figure}

The template adopted in this study is shown in Figure~\ref{fig:BLM-template}. BLM templates are typically constructed around three core elements: external rules \textit{E}, internal rules \textit{I} and relational operators \textit{R} \citep{merlo2023blackbirdlanguagematricesblm}.
The external rule \textit{E} of the phenomenon that distinguishes passive from active voice concerns verbal morphology.  
The internal rule \textit{I} relates to the syntax–semantic mapping of the verb’s arguments. In passive constructions, the \textcolor{blue}{Theme} is promoted to syntactic subject position and agrees with the verb.  
The \textit{R} components specifies how the linguistic objects of the BLM are connected within the template. In the proposed template, \textit{R} corresponds to sentence type (\textit{R\textsubscript{ST})} and number of arguments superficially realized in the sentence (\textit{R\textsubscript{Arg}}). These two linguistic phenomena are external to the one under investigation and do not influence the learning of \textit{E} and \textit{I} rules. We explore the contrast between interrogative and declarative sentences, but also between bi-argumental and mono-argumental structures. 

The inclusion of both declarative and interrogative clauses is partly motivated by the structure of the available treebanks and the distributions of sentences across treebanks. For example, the French QuestionBank (FQB, \citealt{SeddahCandito2016}) exclusively contains interrogative sentences, and allows the retrieval of a large number of instances for task training and testing purposes. Declaratives are also present in large scale in the UD treebanks of French and Italian \citep{SamoMerlo2021}.  From a linguistic perspective, sentence type constitutes an independent dimension with respect to passive voice alternation. Interrogative marking may affect word order, but does not interfere with the morphological marking of passive voice (\textit{E}), nor with the argument-structure reconfiguration involved in the promotion of the \textcolor{blue}{Theme} (\textit{I}). 

We therefore develop a template that manipulates the organization of the three core elements \textit{E}, \textit{I}, and \textit{R}. Elements of \textit{E} include the verb inflected for the active voice (\textcolor{teal}{Vact}) and for the passive voice (\textcolor{orange}{Vpass}). The \textit{I} component concerns the syntactic functions of the core arguments,  \textcolor{violet}{Agents} and \textcolor{blue}{Themes}. The \textit{R} component specifies whether both \textcolor{violet}{Agents} and \textcolor{blue}{Themes} are present in the same string (or only one of these arguments appears, \textit{R\textsubscript{Arg}}) and whether the sentence is interrogative or declarative. Interrogative sentences are marked typographically by the interrogative symbol \textcolor{red}{?}, and also by the presence of \textit{wh}-elements (\textit{R\textsubscript{ST})}.

\paragraph{Context Set}

Formally, the context set instantiates a 2 × 2 × 2 design crossing voice (active/passive), number of argument realizations (one or two overt arguments), and sentence type (interrogative vs. declarative). The eighth sentence is left uninstantiated. The organization of the context reflects two independent alternations built on the different \textit{R}s (\textit{R}\textsubscript{Arg} and \textit{R}\textsubscript{ST}). 

Each pair of consecutive sentences forms a minimal interrogative–declarative paradigm: in all four pairs, the first member is interrogative and the second is declarative. For every four sentences, the first two show two overt arguments, while the last two show only one argument. The resulting structure creates a double alternation pattern (and a triple one with active-passive): sentence type alternates within each pair, while voice and argument realization alternate across pairs. 

The first pair (1–2) consists of active clauses with two overt arguments. The second pair (3–4) maintains the active voice, but in mono-argumental structures. The third pair (5–6) shifts to the passive voice while restoring the bi-argumental configuration, with the \textcolor{blue}{Theme} promoted to subject position and the \textcolor{violet}{Agent} realized as an overt agentive complement. Finally, the fourth pair is designed to instantiate mono-argumental passive clauses; however, only the interrogative configuration (7) is provided in the context. The corresponding declarative passive without an overt agent (8) is omitted and must be inferred. The missing final sentence can therefore be recovered only by integrating all the patterns simultaneously. 


\paragraph{Answer Set}

The correct answer is the declarative passive sentence without an agentive prepositional phrase. 
The answer set is designed to cover all combinations of argument number (one or two arguments) and voice (active vs. passive), and an additional interrogative option. Therefore, the errors are of three types: (i) errors of \textsc{Voice} when the verb is inflected in the active voice, violating \textit{E} and implicitly \textit{I}; (ii) errors of \textsc{Number of arguments} when there are two arguments; (iii) errors of \textsc{Sentence type} when the sentence is not a declarative, both violations of \textit{R}.
In our case, candidate answer 2 (\textcolor{blue}{Th} \textcolor{orange}{Vpass} \textcolor{violet}{Ag}) is an error of \textsc{Number of arguments} because it also shows an additional argument, the agentive PP; sentence 3 (\textcolor{violet}{Ag} \textcolor{teal}{Vact}) is an error of \textsc{Voice} because it is a mono-argumental declarative, but not a passive structure; sentence 4 (\textcolor{violet}{Ag} \textcolor{teal}{Vact} \textcolor{blue}{Th}) is an error of both \textsc{Voice} and \textsc{Number of Arguments}; sentence 5 (\textcolor{red}{question}) is mainly an error of \textsc{Sentence Type}.
%

\section{Natural and Synthetic Data for BLM}
\label{data}

The template illustrated in Figure \ref{fig:BLM-template} is instantiated with natural and synthetic data. Examples are given in French and Italian respectively in Figure \ref{fig:BLM-template-FR} and Figure \ref{fig:BLM-template-IT}.\footnote{Data is available at \url{https://www.idiap.ch/
en/scientific-research/data/blm-passF} and
\url{https://www.idiap.ch/en/scientific-research/
data/blm-passI}.}

\subsection{Natural data}
\label{naturaldata}

\begin{figure*}
  \centering
  \scriptsize
  \begin{tabular}{c p{0.55\textwidth} p{0.35\textwidth}}
    \hline
    & \textbf{Natural} &  \textbf{Synthetic} \\ \hline
    \multicolumn{3}{c}{\textsc{Context Set}}  \\\hline
    \hline
    1  & \textit{\textcolor{blue}{Que} \textcolor{teal}{signifie} \textcolor{violet}{l'acronyme NASA} \textcolor{red}{?}}	& \textit{\textcolor{violet}{Le garçon} \textcolor{teal}{jette}-t-\textcolor{violet}{il} \textcolor{blue}{la pierre}\textcolor{red}{?}}\\
    &  `what does the acronym NASA mean?' (FQB$\_$TREC-fr-584) & 	`Does the boy throw the stone?'\\
    2 &  \textit{\textcolor{violet}{Leur album No Dice} \textcolor{teal}{obtient} \textcolor{blue}{un excellent accueil.}}  & \textit{\textcolor{violet}{L'équipe} \textcolor{teal}{félicite} \textcolor{blue}{le gagnant}.}  \\
    & `Their album \textit{No Dice} obtained an excellent reception' (fr-ud-train$\_$06225) & `The team congratulates the winner' \\
    3  & \textit{Combien de temps \textcolor{teal}{dure} \textcolor{violet}{le voyage entre Tokyo et Niigata} \textcolor{red}{?} }& \textit{\textcolor{violet}{L'équipe} \textcolor{teal}{coûte}-t-\textcolor{violet}{elle}\textcolor{red}{?}} \\
    & `How long does the journey from Tokyo to Niigata take?' (FQB$\_$TREC-fr-111) &  `Does the team cost?' \\ 
    4  &  \textit{\textcolor{violet}{Bryon Anthony McCane} \textcolor{teal}{naît} en 1976 d'une mère italienne et d'un père afro-américain.} & \textit{\textcolor{violet}{Le chanteur} \textcolor{teal}{chante}.}	\\ & `Bryon Anthony McCane was born in 1976 to an Italian mother and an African-American father.'	       (fr-ud-train$\_$00556)   & `The singer sings' \\
    5  &    \textit{\textcolor{blue}{Les moteurs rotatifs} \textcolor{orange}{étaient produits} \textcolor{violet}{par qui} \textcolor{red}{?}	 }     & \textit{Comment \textcolor{blue}{la scène} \textcolor{orange}{est}-\textcolor{blue}{elle} \textcolor{orange}{décrite} \textcolor{violet}{par l'écrivain} \textcolor{red}{?}} \\
    & `By whom were rotating engines invented?' (FQB$\_$TREC-fr-886) & `How is the scene described by the writer?'
    \\    6 &      \textit{\textcolor{blue}{Uber} \textcolor{orange}{a été fondé} \textcolor{violet}{par Garrett Camp, Travis Kalanick et Oscar Salazar} en 2009.}   & \textit{\textcolor{blue}{Un livre} \textcolor{orange}{est écrit} \textcolor{violet}{par l'auteur}.} \\ & `Uber was founded by Garrett Camp, Travis Kalanick and Oscar Salazar in 2009' (fr-ud-train$\_$02964) & `A book is written by the author' \\
    7 & \textit{Dans quelle province française \textcolor{blue}{le cognac } \textcolor{orange}{est}-\textcolor{blue}{il} \textcolor{orange}{produit}} \textcolor{red}{?} & \textit{Quand \textcolor{blue}{la musique} \textcolor{orange}{a}-t-\textcolor{blue}{elle} \textcolor{orange}{été composée}\textcolor{red}{?}}\\
    & `In which French province is cognac produced?' (FQB$\_$TREC-fr-1134) & `When was the music composed?'\\
    8 & ??? & ???\\
    \hline
    \multicolumn{3}{c}{\textsc{Answer Set}}  \\\hline
    1  & \textit{\textbf{\textcolor{blue}{Le château} \textcolor{orange}{est} ensuite \textcolor{orange}{vendu} plusieurs fois }} & \textit{\textbf{\textcolor{blue}{Les données} \textcolor{orange}{ont été analysées}.}}\\
    & `The castle is then sold several times' (fr-ud-train$\_$00010) & The data were analyzed \\
    2 & \textit{\textcolor{blue}{Les travaux} \textcolor{orange}{furent dirigés} \textcolor{violet}{par le Florentin Girolamo della Robbia et les Tourangeaux Pierre Gadier et Gatien François.}} & \textit{\textcolor{blue}{La langue} \textcolor{orange}{est apprise} par \textcolor{violet}{l'étudiant}.}\\ & 
    `The works were directed by the Florentine Girolamo della Robbia and the Turingians Pierre Gadier and Gatien François' (fr-ud-train$\_$07752) & `The language is learned by the student.'\\
   3  & En tout cas \textcolor{violet}{je} n'y \textcolor{teal}{retourne} pas et ma collègue non plus ! & \textit{\textcolor{violet}{Le programmeur} \textcolor{teal}{code}}. \\  & `In any case, I’m not going back there and neither my colleague!' (fr-ud-train$\_$00110) & `The programmer codes.'  \\
   4  & \textit{En mars 2010, \textcolor{violet}{il} \textcolor{teal}{signe} \textcolor{blue}{son premier contrat professionnel} avec Birmingham City.} & \textit{\textcolor{violet}{Le parent } \textcolor{teal}{paie} \textcolor{blue}{la facture}}. \\ &   `In March 2010, he signed his first professional contract with Birmingham City'  (fr-ud-train$\_$01336) & `The parent pays the bill' \\ 
   5 & \textit{\textcolor{blue}{Quel âge} \textcolor{teal}{a} \textcolor{violet}{le Soleil} \textcolor{red}{?}} & \textit{Quand \textcolor{blue}{le colis} \textcolor{orange}{a}-t-\textcolor{blue}{il} \textcolor{orange}{été reçu}\textcolor{red}{?}}  \\
   & `How old is the Sun?'(FQB$\_$TREC-fr-334) & `When was the package received?' \\

  \end{tabular}
  
  \caption{Natural and synthetic examples in French, glosses and ID number (natural data) for reference within the explored treebanks. We have coloured code the core elements of the BLM-template. Correct answer in bold.}
  \label{fig:BLM-template-FR}
\end{figure*}

\begin{figure*}
  \centering
  \scriptsize
   \begin{tabular}{c p{0.55\textwidth} p{0.35\textwidth}}
    \hline
    & \textbf{Example Natural} &  \textbf{Example Synthetic} \\ \hline
    \multicolumn{2}{c}{\textsc{Context Set}}  \\\hline
    \hline
    1 & \textit{Quando \textcolor{violet}{Innsbruck} \textcolor{teal}{ospitò} \textcolor{blue}{le Olimpiadi Invernali}\textcolor{red}{?} }
      & \textit{\textcolor{violet}{Lo scrittore} \textcolor{teal}{finisce} \textcolor{blue}{il romanzo}\textcolor{red}{?}} \\
      & `When did Innsbruck host the Winter Olympic Games?' (ISDT, quest-991) 
      & `Does the writer finish the novel?' \\

    2 & \textit{\textcolor{violet}{I carabinieri} gli \textcolor{teal}{hanno recapitato} \textcolor{blue}{il decreto di revoca degli arresti domiciliari.} }
      & \textit{\textcolor{violet}{Il giocatore} \textcolor{teal}{segna} \textcolor{blue}{un gol}.}  \\
      & `The Carabinieri delivered to him the order revoking his house arrest.' (ISDT, isst$\_$tanl-280)
      & `The player scores a goal.' \\

    3 & \textit{A che velocità \textcolor{teal}{viaggia} \textcolor{violet}{la luce}\textcolor{red}{?}} 
      & \textit{\textcolor{violet}{Il bambino} \textcolor{teal}{disegna}\textcolor{red}{?}} \\
      & `At what speed does light travel?' (ISDT, quest-656)
      & `Does the child draw?' \\

    4 & \textit{E nel frattempo \textcolor{violet}{politici, ambasciatori e industriali} \textcolor{teal}{discutevano} di petrolio.} 
      & \textit{\textcolor{violet}{L'editore} \textcolor{teal}{corregge}.} \\ 
      & `And in the meantime politicians, ambassadors, and industrialists were discussing oil.' (ISDT, isst$\_$tanl-186) & `The publisher corrects.' \\
      
    5 & \textit{\textcolor{blue}{Chi} \textcolor{orange}{è stato sconfitto} \textcolor{violet}{da Andrei Medvedev} nella finale del Torneo di Monte Carlo \textcolor{red}{?}} 
      & \textit{Quando \textcolor{orange}{sono annaffiate} \textcolor{blue}{le piante} \textcolor{violet}{dal giardiniere}\textcolor{red}{?}} \\     
      & `Who was defeated by Andrei Medvedev in the Monte Carlo tournament final?' (ISDT, quest-553)       & `When are the plants watered by the gardener?'\\

    6 & \textit{\textcolor{blue}{Le leggi} \textcolor{orange}{sono promulgate} \textcolor{violet}{dal Presidente della Repubblica} entro un mese dall' approvazione} 
      & \textit{\textcolor{blue}{La squadra} \textcolor{orange}{è tifata} \textcolor{violet}{dal bambino}.} \\ 
      & `The laws are promulgated by the President of the Republic within one month of approval.' (ISDT, tut-1546) & `The team is supported by the child.' \\

    7 & \textit{Come \textcolor{orange}{può essere eliminato} \textcolor{blue}{il rigetto del sistema immunitario}\textcolor{red}{?}} 
      & \textit{Quando \textcolor{orange}{è stato sollevato} \textcolor{blue}{il peso}\textcolor{red}{?}} \\
      & `How can immune system rejection be eliminated?' (quest-297)
      & `When was the weight lifted?' \\
8 & ??? \\
    \hline
    \multicolumn{2}{c}{\textsc{Answer Set}}  \\\hline

    1 & \textit{\textbf{Dopo tre anni di discussioni \textcolor{blue}{la proposta} \textcolor{orange}{non venne accolta}.}} 
      & \textbf{\textit{\textcolor{blue}{Lo strumento} \textcolor{orange}{è stato suonato}.}} \\
      & `After three years of discussions, the proposal was not accepted.' (VIT-1304)
      & `The instrument was played.' \\ 

    2 & \textit{\textcolor{blue}{La lettera di trasporto aereo} \textcolor{orange}{viene emessa }\textcolor{violet}{dal mittente} in tre esemplari originali.} 
      & \textit{\textcolor{blue}{L'email} \textcolor{orange}{è inviata} \textcolor{violet}{dal capo}.} \\
      & `The air waybill is issued by the sender in three original copies.' (ISDT, splet-161) 
      & `The email is sent by the boss.' \\ 

    3 & \textit{\textcolor{violet}{La squadra nazionale italiana di calcio} \textcolor{teal}{giocherà} \textcolor{blue}{la prima partita del Campionato mondiale} il 12 giugno, ad Hannover, con il Ghana.} 
      & \textit{\textcolor{violet}{Il cassiere} \textcolor{teal}{conta} \textcolor{blue}{i soldi}.} \\
      & `The Italian national football team will play its first World Cup match on June 12 in Hanover against Ghana.' (ISDT, 2Parole$\_$1-214)  & `The cashier counts the money.' \\

    4 & \textit{\textcolor{violet}{I disegni} \textcolor{teal}{risentono} delle influenze arabe e a volte anche di quelle francesi}  
      & \textit{\textcolor{violet}{Lo chef} \textcolor{teal}{cucina}.} \\
      & `The drawings show Arab influences and sometimes French ones as well.' (ISDT, isst$\_$tanl-3212)
      & `The chef cooks.' \\ 

    5 & \textit{\textcolor{violet}{Chi} \textcolor{teal}{possiede} \textcolor{blue}{la Saab}\textcolor{red}{?} }
      & \textit{\textcolor{violet}{L'autista} \textcolor{teal}{parcheggia} \textcolor{blue}{l'auto}\textcolor{red}{?}}  \\
      & `Who owns the Saab?' (ISDT, quest-422) 
      & `The driver parks the car?' \\
  \end{tabular}
    \caption{Natural and synthetic examples in Italian, glosses and ID number (natural data) for reference within the explored treebanks. We have coloured code the core elements of the BLM-template.  Correct answer in bold.}
  \label{fig:BLM-template-IT}
\end{figure*}

All the sentences are extracted from UD treebanks. We retrieved natural occurring sentences with the system \textsc{GrewMatch (match.grew.fr)} \citep{guillaume-2021-graph}. For French, we extracted questions from the French QuestionBank v.2.17 (FQB, 23,345 tokens; \citealt{SeddahCandito2016}) containing only questions from news and nonfiction sources.  We extracted declarative sentences from the French GSD data v.2.17 (389,364 tokens;   \citealt{GuillaumeDeMarneffePerrier2019}), which contains data from blogs, news, reviews and encyclopeadic entries. For Italian, we retrieved sentences from ISDT v.2.17 (278,424 tokens; \citealt{BoscoEtAl2014Evalita}) containing data from legal texts, news and Wikipedia and VIT v.2.17 (259,625 tokens; \citealt{AlfieriTamburini2016}), which which consists of news and nonfiction texts. 

The sentence types of the BLM template were directly instantiated using sentences that matched the corresponding structures in the treebanks. Each context and answer sentence was retrieved using simple queries. The \textcolor{violet}{Agent} in the sentences was identified as the target of the \texttt{nsubj} dependency associated with a variable V in active sentences, and as the dependent of a \texttt{obl:agent} dependency in passive sentences. The \textcolor{blue}{Theme} was identified as the dependent of \texttt{obj} in active sentences and of the \texttt{nsubj:pass} dependency in passive sentences.
To retrieve questions, we added the query for an interrogative marker corresponding to a variable Q whose \textsc{form} was annotated as "\textcolor{red}{?}". We restricted our search to  sentences -- intended here as the output of an instance of the query -- with a single verb by excluding output sentences containing additional verbs, using the constraint Y (a variable different than V) whose \textsc{upos} was "VERB". 

The combination of these factors led to the retrieval of natural instances for each sentence of the context and the answer set.\footnote{All queries are available in Table \ref{table-queries-app} in the Appendix.}

Samples of the naturally extracted data were manually checked to assess whether the queries returned the relevant structures.

\subsection{Synthetic data}
\label{syntheticdata}
We created our dataset using the 
DeepSeek-V3 model \citep{guo2025deepseekr1}, adopting a hybrid approach that combines LLM generation with explicit linguistic constraints. This ensured the generated sentences remained grammatically well-formed and semantically coherent within each alternation paradigm. Specifically, the prompt instructed the conversationalAI to generate a structured linguistic dataset consisting of ``100 simple sentences with frequent subjects, verbs and objects'" (5 interactions) in English and transform them into eight distinct syntactic variants corresponding to the context and answer set structures (interrogative, intransitive, and passive forms with and without agentive phrases), with specific formatting requirements for direct import into spreadsheet software. In later interactions, the model was required to translate the sentences into French and Italian. A native speaker of Italian and proficient in French and English, manually checked the translations. The intervention was minimal, as the sentences were generally well-formed.\footnote{Out of 4,000 sentences (500 for each of the eight structures), only 43 were manually corrected for French (0.01) and 64 for Italian (0.02). The intervention primarily ensured that sentences remained unambiguously transitive or intransitive, removing verbs with reflexive-like elements when relevant for the template.}

English was adopted as a pivot language \citep{wendler-etal-2024-llamas}  to construct a single abstract dataset before its language-specific realization. This design choice ensures that the French and Italian versions do not stand in a derivational relationship to each other, but instead instantiate the same underlying synthetic templates. As a result, the synthetic status of the data is methodologically consistent across languages: neither dataset is more natural, or derived, than the other, since both originate from the same controlled generative procedure.

The sentences were specifically designed to isolate the factors under investigation: as shown in Figures \ref{fig:BLM-template-FR} and \ref{fig:BLM-template-IT}, they included exclusively the elements relevant to the BLM (\textcolor{violet}{agents}, \textcolor{blue}{themes}, verbs in \textcolor{teal}{active} and \textcolor{orange}{passive} voice and \textcolor{red}{interrogative markers}).

\section{Experiments}
\label{experiments}
The experiments systematically compare natural and synthetic data in a highly structured setup, the BLM test suite. The goal is to evaluate how much each type of data helps language models identify and abstract the core linguistic elements underlying the passive alternation. We explore the behavior of a simple probe through
a series of experiments.

We adopt a feed-forward neural network (FFNN) architecture as described in \citet{samo-etal-2023-blm}. For each sentence in the BLM, we compute an embedding by averaging its token representations obtained from pretrained models. The FFNN takes as input the concatenated embeddings representing the context, is trained using a max-margin loss objective, and predicts the answer whose embedding achieves the highest cosine similarity with the network’s output. Following previous work \citep{nastase-etal-2025-multilingual}, we first test embeddings from monolingual ELECTRA models (French: \textit{dbmdz/electra-base-french-europeana-cased-discriminator}; Italian: \textit{dbmdz/electra-base-italian-xxl-cased-discriminator}) and then from a multilingual model (\textit{google/electra-base-discriminator}). Monolingual and multilingual models may exhibit asymmetries in performance due to differences in their token representations \citep{samo2026modellingmorphologyverbalparadigms}.

We run both experiments when training and testing belong to  the same data set type and also experiments that cross the two types. Each dataset, constituted of 2000 BLM instances, is split into separate training and testing sets (80\%–20\% split), ensuring no overlap of instances between them to control data leakage. 

These settings are referred to as \textsc{SynSyn} (train and test on synthetic) and \textsc{NatNat} (train and test on natural). \textsc{SynNat} refers to training the model with synthetic data and testing its ability to generalize to natural instances, while \textsc{NatSyn} denotes training on BLM-templates instantiated with natural data and testing on its ability to abstract to synthetic instances.

\subsection{Results: Monolingual}\

\begin{figure}
    \centering
    \includegraphics[width=1\linewidth]{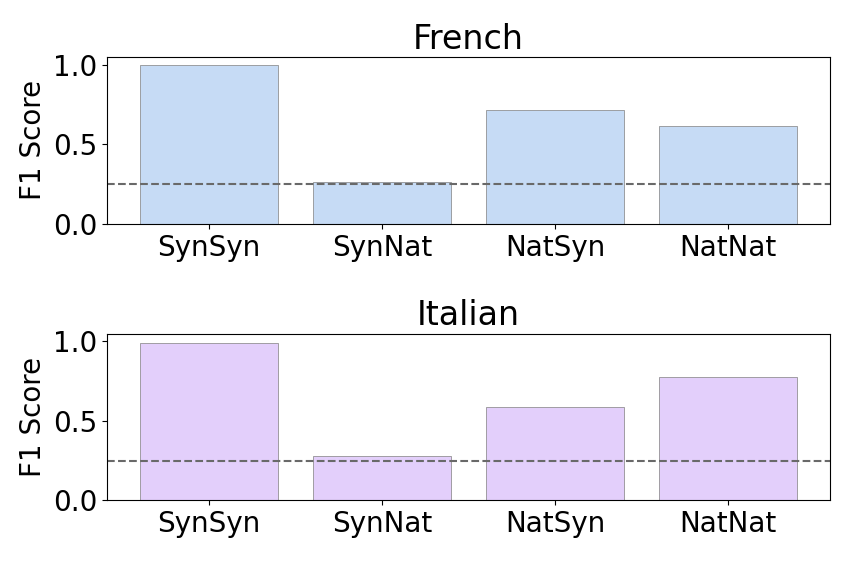}
    \caption{F1 scores across training and test suites in monolingual models. The grey dotted line indicates chance level.}
    \label{fig:F1-monolingual}
\end{figure}

Figure \ref{fig:F1-monolingual} shows the results, measured in F1 scores, for the monolingual models. Among the monolingual models, the full synthetic suite achieves the highest performance (French 1.00; Italian 0.99), outperforming the natural suites (French 0.62; Italian 0.77).

When models are trained on synthetic data but tested on natural data, performance drops significantly, reaching near-chance levels (French 0.29; Italian 0.28). In contrast, training on natural data and testing on synthetic data yields performance comparable to that of the full natural suites, with French even showing improved performance with respect to natural test suites.



\begin{figure}
    \centering
    \includegraphics[width=1\linewidth]{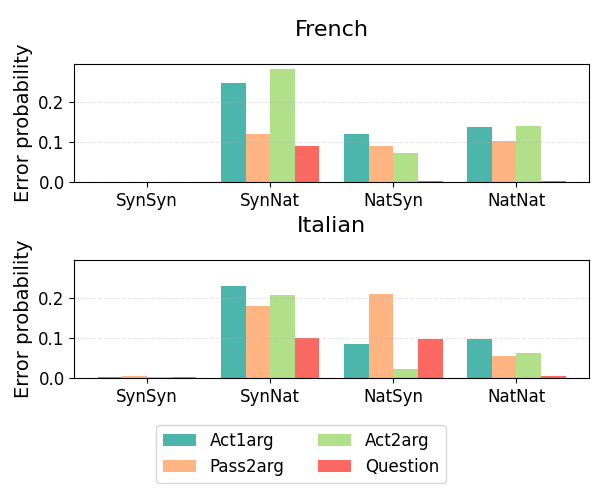}
    \caption{Error probabilities and types of errors (indicated by the selected erroneous answer) in monolingual models. Act = Active; Pass = Passive; arg = arguments.}
    \label{fig:error-analysis-monolingual}
\end{figure}

The error analysis is presented in Figure \ref{fig:error-analysis-monolingual}. Across all suites, errors related to \textsc{Sentence Type} (questions) are never the most prominent while the other errors are more evenly distributed. However, in the Italian \textsc{NatSyn}, the most prominent error type involves passive structures with two arguments, suggesting that voice was a learned feature.

\subsection{Comparing Monolingual and Multilingual Embeddings}

\begin{figure}
    \centering
    \includegraphics[width=1\linewidth]{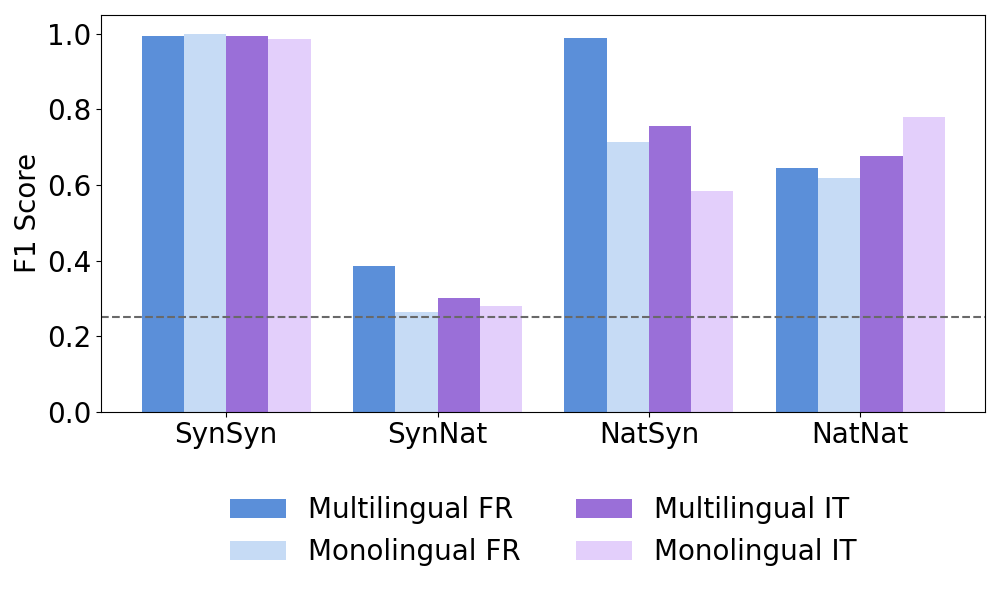}
    \caption{F1 scores across conditions, languages and models. The gray dotted line indicates chance level. IT = Italian, FR = French.}
    \label{fig:F1-scores-Experiment1}
\end{figure}

\begin{figure}
    \centering
    \includegraphics[width=1\linewidth]{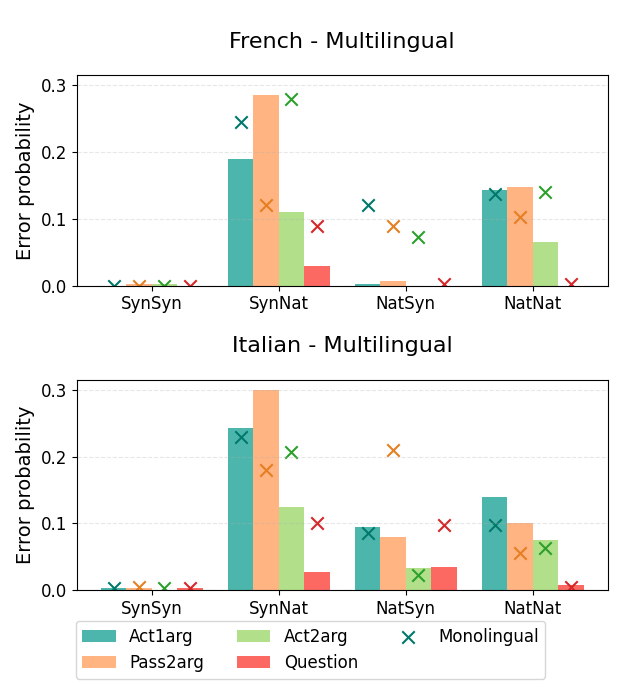}
    \caption{Error probabilities and types of errors (indicated by the selected erroneous answer) in multilingual models. Crosses repeat values of error bars in monolingual model for comparison (see Figure \ref{fig:error-analysis-monolingual}). Act=Active, Pass=Passive, arg=arguments.}
    \label{fig:mono-vs-multi}
\end{figure}

Figure \ref{fig:mono-vs-multi} shows the results of the comparison between monolingual and multilingual models. The results follow the same pattern across languages and models. Multilingual models show competitive, and in some cases superior peformance. In particular, the multilingual model reaches ceiling performance in the \textsc{NatSyn} suite  for French (0.99), indicating strong comprehension of the BLM and generalisation ability from natural to synthetic instances.

Consider the error analysis. In the \textsc{NatNat} setting the most frequent incorrect responses in the multilingual models involve \textsc{Voice} or \textsc{Number of Arguments}. This pattern indicates that active sentences with two arguments (i.e., violations of two rules) are not considered plausible answers. Errors related to \textsc{Sentence Type} are minimal in the multilingual models, even in the lowest-performing configuration. This suggests that the models consistently learn the sentence type encoded in the BLM template across languages and training conditions.

However, multilingual and monolingual models differ in the type of error that is most prominent in the lowest-performing setting, \textsc{SynNat}. In both languages, the most frequent errors produced by the multilingual models are passive structures, suggesting that these models successfully learn verb voice from synthetic training data. In contrast, monolingual models show a higher proportion of active constructions as their most prominent error.

\section{Discussion}
\label{discussion}
Although the full synthetic suites outperform the full natural suites, overall performance drops significantly when models are trained on synthetic data but tested on natural data (aggregated, \textit{t}-test: \textit{t}(6) = 27.07, \textit{SE} = 0.025). Although models trained on natural data show lower absolute performance, their results are more stable across different evaluation settings.

Despite the fact that models trained and tested on synthetic data achieve near-ceiling performance, this competence does not reliably generalize the learnt structure to natural data. Training on synthetic data may bias the model toward artifacts introduced by the synthetic constructions themselves. As a consequence, models fail to generalize to natural test suites, which are structurally and contextually more complex—where complexity refers to the presence of additional constituents that are not directly relevant to the task. In contrast, models trained on natural data show more robust cross-condition generalization, performing competitively not only on natural test suites but also on synthetic ones.

Natural sentences—here extracted from UD treebanks—exhibit greater lexical variability (e.g. proper nouns or hapaxes) and additional linguistic objects surrounding the target syntactic configuration. Rather than hindering learning, however, this variability appears to promote abstraction of the core elements of the task, although not reaching ceiling performance in the testing suites.



The Universal Dependencies (UD) framework --as a particularly curated and cross-linguistically valid example of structured and annotated data in general-- plays a central methodological role in this study. By extracting instances through dependency relations and morphological features, we were able to construct datasets that are both naturally occurring and structurally controlled. Crucially UD allows the same query to be executed consistently across different languages and text genres. This makes it possible to retrieve comparable structures in a uniform way, without redefining language-specific search criteria each time. In this sense, UD and structured annotations in general  function as a bridge between raw language data and the need to build theoretically motivated,  structured diagnostics paradigms for natural data.

\section{Related Work}
\label{related}
Language models have demonstrated good performance on lexically restricted verb alternations and thematic roles, especially in English \citep{kann-etal-2019-verb, warstadt-etal-2019-investigating, wilson2023abstract, samo-etal-2023-blm, proietti-etal-2022-bert}. Our work expands computational approaches to the passive alternation \citep{sasano-etal-2013-automatic,leong-linzen-2023-language}.

BLMs share objectives with datasets relying on synthetic or carefully controlled materials, designed to target specific grammatical phenomena under controlled conditions, often using minimal pairs or carefully crafted paradigms to probe the linguistic knowledge of large language models across languages \citep{warstadt-etal-2019-investigating, warstadt-etal-2020-blimp-benchmark,xiang-etal-2021-climp,suijkerbuijk2025/COLI_a_00559}.
Unlike datasets of minimal pairs, though, the BLM puzzle presents a complex structure and expands the minimal contrasts to many dimensions of variation.

While synthetic data allow broad and systematic coverage, they can introduce distributional biases, encouraging models to rely on superficial regularities rather than abstract rules \citep{nadas-etal-2025-survey, griffiths2024bayes}. Our results partly resonate with the findings of \citet{zhang2025doestrainingsyntheticdata}, who show that models trained on synthetic data may continue to rely on superficial heuristics that generalize poorly to targeted evaluations, even when overall benchmark performance improves. 


Natural data drawn from annotated and structured resources provide rich linguistic information at the word \citep{batsuren-etal-2022-unimorph} or sentence level \citep{nivre2015towards, de2021universal},  offering authentic variation and contextual richness. Such data allow researchers to systematically probe language models, designing tasks that implicitly target complex grammatical phenomena, including agreement \citep{jumelet-etal-2025-multiblimp} and argument structure alternations in morphologically-rich languages \citep{samo2026modellingmorphologyverbalparadigms}. By bridging naturalistic data and controlled evaluation, structured corpora provide a principled framework for investigating the inner workings of language models.


\section{Conclusion}
\label{conclusion}
Our study shows that while synthetic data allow models to achieve near-perfect performance on in-distribution tasks, they fail to support robust generalization to natural language. In contrast, models trained on natural sentences, systematically retrieved from Universal Dependencies, capture the underlying patterns of passive alternation and generalize effectively across both natural and synthetic test sets. These findings highlight the enduring value of authentic, linguistically-grounded data for evaluating large language models: variability and complexity in natural corpora encourage models to learn abstract structural generalizations rather than memorizing surface patterns. Structured and annotated natural data remain crucial for rigorous linguistic evaluation and for probing the true syntactic and semantic knowledge of pretrained models cross-linguistically.

\section*{Limitations}

Future work could address the limitations of this contribution by expanding language coverage and alternation phenomena, exploring additional models and architectures, and performing comprehensive validation, as well a human upperbound.

\section*{Ethics}

We used datasets derived from publicly available corpora, which may include content such as news articles and other publicly accessible materials. It is important to note that these datasets may contain sensitive or potentially upsetting topics. We acknowledge that such content could be distressing to some individuals. We encourage users to approach the results with awareness of these considerations.








\section{Bibliographical References}\label{sec:reference}

\bibliographystyle{lrec2026-natbib}
\bibliography{lrec2026-example}

@inproceedings{leong-linzen-2023-language,
    title = "Language Models Can Learn Exceptions to Syntactic Rules",
    author = "Leong, Cara Su-Yi  and
      Linzen, Tal",
    editor = "Hunter, Tim  and
      Prickett, Brandon",
    booktitle = "Proceedings of the Society for Computation in Linguistics 2023",
    month = jun,
    year = "2023",
    address = "Amherst, MA",
    publisher = "Association for Computational Linguistics",
    url = "https://aclanthology.org/2023.scil-1.11/",
    pages = "133--144"
}

@inproceedings{AlfieriTamburini2016,
  author    = {Alfieri, Lorenzo and Tamburini, Fabio},
  title     = {{(Almost) Automatic Conversion of the Venice Italian Treebank into the Merged Italian Dependency Treebank Format}},
  booktitle = {Proceedings of the Third Italian Conference on Computational Linguistics (CLiC-IT 2016)},
  year      = {2016},
  address   = {Napoli, Italy},
  month     = {dec},
  pages     = {19--23}
}

@inproceedings{BoscoEtAl2014Evalita,
  author    = {Bosco, Cristina and Dell'Orletta, Felice and Montemagni, Simonetta and Sanguinetti, Manuela and Simi, Maria},
  title     = {{The Evalita 2014 Dependency Parsing Task}},
  booktitle = {Proceedings of CLiC-it 2014 and EVALITA 2014},
  year      = {2014},
  publisher = {Pisa University Press},
  address   = {Pisa, Italy},
  pages     = {1--8},
  isbn      = {978-886741-472-7}
}

@article{GuillaumeDeMarneffePerrier2019,
  author  = {Guillaume, Bruno and de Marneffe, Marie-Catherine and Perrier, Guy},
  title   = {Conversion et am{\'e}liorations de corpus du fran{\c{c}}ais annot{\'e}s en Universal Dependencies},
  journal = {Traitement Automatique des Langues},
  year    = {2019},
  volume  = {60},
  number  = {2},
  pages   = {71--95},
  note    = {HAL: hal-02267418}
}

@inproceedings{SeddahCandito2016,
  author    = {Seddah, Djam{\'e} and Candito, Marie},
  title     = {Hard Time Parsing Questions: Building a QuestionBank for French},
  booktitle = {Proceedings of the Tenth International Conference on Language Resources and Evaluation (LREC 2016)},
  year      = {2016},
  month     = {may},
  address   = {Portoro{\v{z}}, Slovenia}
}

@incollection{wals-107,
  author    = {Anna Siewierska},
  booktitle = {The World Atlas of Language Structures Online},
  editor    = {Matthew S. Dryer and Martin Haspelmath},
  publisher = {Zenodo},
  title     = {Passive Constructions (v2020.4)},
  type      = {Data set},
  url       = {https://doi.org/10.5281/zenodo.13950591},
  year      = {2013},
  doi       = {10.5281/zenodo.13950591}
}

@book{chomsky1965,
  title={Aspects of the Theory of Syntax},
  author={Chomsky, Noam},
  year={1965},
  publisher={MIT Press},
address = {Cambridge, MA}
}

@article{gibson_fedorenko_2013,
  title   = {{The Need for Quantitative Methods in Syntax and Semantics Research}},
  author  = {Gibson, Edward and Fedorenko, Evelina},
  journal = {Language and Cognitive Processes},
  volume  = {28},
  number  = {1--2},
  pages   = {88--124},
  year    = {2013},
  doi     = {10.1080/01690965.2010.515080}
}

@article{GoyalMahmoud2024,
  author    = {Anita Goyal and Samira Mahmoud},
  title     = {{A Systematic Review of Synthetic Data Generation Techniques for NLP}},
  journal   = {Electronics},
  volume    = {13},
  number    = {17},
  pages     = {3509},
  year      = {2024},
  publisher = {MDPI},
  url = {https://doi.org/10.3390/electronics13173509}
}

@article{nissim2025challengingabilitieslargelanguage,
      title={{Challenging the Abilities of Large Language Models in Italian: a Community Initiative}}, 
      author={Malvina Nissim and Danilo Croce and Viviana Patti and Pierpaolo Basile and Giuseppe Attanasio and Elio Musacchio and Matteo Rinaldi and Federico Borazio and Maria Francis and Jacopo Gili and Daniel Scalena and Begoña Altuna and Ekhi Azurmendi and Valerio Basile and Luisa Bentivogli and Arianna Bisazza and Marianna Bolognesi and Dominique Brunato and Tommaso Caselli and Silvia Casola and Maria Cassese and Mauro Cettolo and Claudia Collacciani and Leonardo De Cosmo and Maria Pia Di Buono and Andrea Esuli and Julen Etxaniz and Chiara Ferrando and Alessia Fidelangeli and Simona Frenda and Achille Fusco and Marco Gaido and Andrea Galassi and Federico Galli and Luca Giordano and Mattia Goffetti and Itziar Gonzalez-Dios and Lorenzo Gregori and Giulia Grundler and Sandro Iannaccone and Chunyang Jiang and Moreno La Quatra and Francesca Lagioia and Soda Marem Lo and Marco Madeddu and Bernardo Magnini and Raffaele Manna and Fabio Mercorio and Paola Merlo and Arianna Muti and Vivi Nastase and Matteo Negri and Dario Onorati and Elena Palmieri and Sara Papi and Lucia Passaro and Giulia Pensa and Andrea Piergentili and Daniele Potertì and Giovanni Puccetti and Federico Ranaldi and Leonardo Ranaldi and Andrea Amelio Ravelli and Martina Rosola and Elena Sofia Ruzzetti and Giuseppe Samo and Andrea Santilli and Piera Santin and Gabriele Sarti and Giovanni Sartor and Beatrice Savoldi and Antonio Serino and Andrea Seveso and Lucia Siciliani and Paolo Torroni and Rossella Varvara and Andrea Zaninello and Asya Zanollo and Fabio Massimo Zanzotto and Kamyar Zeinalipour and Andrea Zugarini},
      year={2025},
      journal={arXiv preprint},
      volume={arXiv: 2512.04759},
      primaryClass={cs.CL},
      url={https://arxiv.org/abs/2512.04759}, 
}

@article{warstadt-etal-2020-blimp-benchmark,
    title = "{BL}i{MP}: The Benchmark of Linguistic Minimal Pairs for {E}nglish",
    author = "Warstadt, Alex  and
      Parrish, Alicia  and
      Liu, Haokun  and
      Mohananey, Anhad  and
      Peng, Wei  and
      Wang, Sheng-Fu  and
      Bowman, Samuel R.",
    editor = "Johnson, Mark  and
      Roark, Brian  and
      Nenkova, Ani",
    journal = "Transactions of the Association for Computational Linguistics",
    volume = "8",
    year = "2020",
    address = "Cambridge, MA",
    publisher = "MIT Press",
    url = "https://aclanthology.org/2020.tacl-1.25/",
    doi = "10.1162/tacl_a_00321",
    pages = "377--392"
}

@inproceedings{batsuren-etal-2022-unimorph,
    title = "{U}ni{M}orph 4.0: {U}niversal {M}orphology",
    author = "Batsuren, Khuyagbaatar  and
      Goldman, Omer  and
      Khalifa, Salam  and
      Habash, Nizar  and
      Kiera{\'s}, Witold  and
      Bella, G{\'a}bor  and
      Leonard, Brian  and
      Nicolai, Garrett  and
      Gorman, Kyle  and
      Ate, Yustinus Ghanggo  and
      Ryskina, Maria  and
      Mielke, Sabrina  and
      Budianskaya, Elena  and
      El-Khaissi, Charbel  and
      Pimentel, Tiago  and
      Gasser, Michael  and
      Lane, William Abbott  and
      Raj, Mohit  and
      Coler, Matt  and
      Samame, Jaime Rafael Montoya  and
      Camaiteri, Delio Siticonatzi  and
      Rojas, Esa{\'u} Zumaeta  and
      L{\'o}pez Francis, Didier  and
      Oncevay, Arturo  and
      L{\'o}pez Bautista, Juan  and
      Villegas, Gema Celeste Silva  and
      Hennigen, Lucas Torroba  and
      Ek, Adam  and
      Guriel, David  and
      Dirix, Peter  and
      Bernardy, Jean-Philippe  and
      Scherbakov, Andrey  and
      Bayyr-ool, Aziyana  and
      Anastasopoulos, Antonios  and
      Zariquiey, Roberto  and
      Sheifer, Karina  and
      Ganieva, Sofya  and
      Cruz, Hilaria  and
      Karah{\'o}ǧa, Ritv{\'a}n  and
      Markantonatou, Stella  and
      Pavlidis, George  and
      Plugaryov, Matvey  and
      Klyachko, Elena  and
      Salehi, Ali  and
      Angulo, Candy  and
      Baxi, Jatayu  and
      Krizhanovsky, Andrew  and
      Krizhanovskaya, Natalia  and
      Salesky, Elizabeth  and
      Vania, Clara  and
      Ivanova, Sardana  and
      White, Jennifer  and
      Maudslay, Rowan Hall  and
      Valvoda, Josef  and
      Zmigrod, Ran  and
      Czarnowska, Paula  and
      Nikkarinen, Irene  and
      Salchak, Aelita  and
      Bhatt, Brijesh  and
      Straughn, Christopher  and
      Liu, Zoey  and
      Washington, Jonathan North  and
      Pinter, Yuval  and
      Ataman, Duygu  and
      Wolinski, Marcin  and
      Suhardijanto, Totok  and
      Yablonskaya, Anna  and
      Stoehr, Niklas  and
      Dolatian, Hossep  and
      Nuriah, Zahroh  and
      Ratan, Shyam  and
      Tyers, Francis M.  and
      Ponti, Edoardo M.  and
      Aiton, Grant  and
      Arora, Aryaman  and
      Hatcher, Richard J.  and
      Kumar, Ritesh  and
      Young, Jeremiah  and
      Rodionova, Daria  and
      Yemelina, Anastasia  and
      Andrushko, Taras  and
      Marchenko, Igor  and
      Mashkovtseva, Polina  and
      Serova, Alexandra  and
      Prud{'}hommeaux, Emily  and
      Nepomniashchaya, Maria  and
      Giunchiglia, Fausto  and
      Chodroff, Eleanor  and
      Hulden, Mans  and
      Silfverberg, Miikka  and
      McCarthy, Arya D.  and
      Yarowsky, David  and
      Cotterell, Ryan  and
      Tsarfaty, Reut  and
      Vylomova, Ekaterina",
    editor = "Calzolari, Nicoletta  and
      B{\'e}chet, Fr{\'e}d{\'e}ric  and
      Blache, Philippe  and
      Choukri, Khalid  and
      Cieri, Christopher  and
      Declerck, Thierry  and
      Goggi, Sara  and
      Isahara, Hitoshi  and
      Maegaard, Bente  and
      Mariani, Joseph  and
      Mazo, H{\'e}l{\`e}ne  and
      Odijk, Jan  and
      Piperidis, Stelios",
    booktitle = "Proceedings of the Thirteenth Language Resources and Evaluation Conference",
    month = jun,
    year = "2022",
    address = "Marseille, France",
    publisher = "European Language Resources Association",
    url = "https://aclanthology.org/2022.lrec-1.89/",
    pages = "840--855"
}

@article{zhang2025doestrainingsyntheticdata,
      title={{Does Training on Synthetic Data Make Models Less Robust?}}, 
      author={Lingze Zhang and Ellie Pavlick},
      year={2025},
      volume={arXiv: 2502.07164},
      journal={arXiv preprint},
      primaryClass={cs.CL},
      url={https://arxiv.org/abs/2502.07164}, 
}

@article{samo2026datasets,
      title={{Datasets for Verb Alternations across Languages: BLM Templates and Data Augmentation Strategies}}, 
      author={Giuseppe Samo and Paola Merlo},
      year={2026},
      volume={arXiv: 2603.15295},
      journal={arXiv},
      primaryClass={cs.CL},
      url={https://arxiv.org/abs/2603.15295}, 
}

@article{nadas-etal-2025-survey,
   title={Synthetic Data Generation Using Large Language Models: Advances in Text and Code},
   volume={13},
   ISSN={2169-3536},
   url={http://dx.doi.org/10.1109/ACCESS.2025.3589503},
   DOI={10.1109/access.2025.3589503},
   journal={IEEE Access},
   publisher={Institute of Electrical and Electronics Engineers (IEEE)},
   author={Nadǎş, Mihai and Dioşan, Laura and Tomescu, Andreea},
   year={2025},
   pages={134615–134633} }

@article{jumelet-etal-2025-multiblimp,
  title = {{MultiBLiMP 1.0: A Massively Multilingual Benchmark of Linguistic Minimal Pairs}} ,
  author = {Jumelet, Jaap and Weissweiler, Leonie and Nivre, Joakim and Bisazza, Arianna},
  journal = {arXiv preprint arXiv:2504.02768},
  year = {2025}
}

@inproceedings{nastase2024exploring1,
    title = "Exploring {I}talian Sentence Embeddings Properties through Multi-tasking",
    author = "Nastase, Vivi  and
      Samo, Giuseppe  and
      Jiang, Chunyang  and
      Merlo, Paola",
    editor = "Dell'Orletta, Felice  and
      Lenci, Alessandro  and
      Montemagni, Simonetta  and
      Sprugnoli, Rachele",
    booktitle = "Proceedings of the Tenth Italian Conference on Computational Linguistics (CLiC-it 2024)",
    month = dec,
    year = "2024",
    address = "Pisa, Italy",
    publisher = "CEUR Workshop Proceedings",
    url = "https://aclanthology.org/2024.clicit-1.70/",
    pages = "620--630",
    ISBN = "979-12-210-7060-6"
}

@inproceedings{nastase2024exploring2,
    title = "Exploring Syntactic Information in Sentence Embeddings through Multilingual Subject-verb Agreement",
    author = "Nastase, Vivi  and
      Samo, Giuseppe  and
      Jiang, Chunyang  and
      Merlo, Paola",
    editor = "Dell'Orletta, Felice  and
      Lenci, Alessandro  and
      Montemagni, Simonetta  and
      Sprugnoli, Rachele",
    booktitle = "Proceedings of the Tenth Italian Conference on Computational Linguistics (CLiC-it 2024)",
    month = dec,
    year = "2024",
    address = "Pisa, Italy",
    publisher = "CEUR Workshop Proceedings",
    url = "https://aclanthology.org/2024.clicit-1.71/",
    pages = "631--643",
    ISBN = "979-12-210-7060-6"
}

@article{rappaporthovav2024variable,
  author    = {Rappaport Hovav, Malka and Levin, Beth},
  title     = {{Variable Agentivity: Polysemy or underspecification}},
  journal   = {Glossa: a journal of general linguistics},
  volume    = {9},
  number    = {1},
  year      = {2024},
  doi       = {10.16995/glossa.10774},
  url       = {https://doi.org/10.16995/glossa.10774}
}

@inproceedings{warstadt-etal-2019-investigating,
    title = {{Investigating BERT's Knowledge of Language: Five Analysis Methods with NPIs}},
    author = "Warstadt, Alex  and
      Cao, Yu  and
      Grosu, Ioana  and
      Peng, Wei  and
      Blix, Hagen  and
      Nie, Yining  and
      Alsop, Anna  and
      Bordia, Shikha  and
      Liu, Haokun  and
      Parrish, Alicia  and
      Wang, Sheng-Fu  and
      Phang, Jason  and
      Mohananey, Anhad  and
      Htut, Phu Mon  and
      Jeretic, Paloma  and
      Bowman, Samuel R.",
    editor = "Inui, Kentaro  and
      Jiang, Jing  and
      Ng, Vincent  and
      Wan, Xiaojun",
    booktitle = "Proceedings of the 2019 Conference on Empirical Methods in Natural Language Processing and the 9th International Joint Conference on Natural Language Processing (EMNLP-IJCNLP)",
    month = nov,
    year = "2019",
    address = "Hong Kong, China",
    publisher = "Association for Computational Linguistics",
    url = "https://aclanthology.org/D19-1286",
    doi = "10.18653/v1/D19-1286",
    pages = "2877--2887",
}

@article{sprouse2017design,
  title={{Design sensitivity and statistical power in acceptability judgment experiments}},
  author={Sprouse, Jon and Almeida, Diogo},
  journal={Glossa: a journal of general linguistics},
  volume={2},
  number={1},
  pages={14},
  doi ={https://doi.org/10.5334/gjgl.236},
  year={2017},
  publisher={Open Library of Humanities}
}

@inproceedings{proietti-etal-2022-bert,
    title = {{Does BERT Recognize an Agent? Modeling Dowty's Proto-Roles with Contextual Embeddings}},
    author = "Proietti, Mattia  and
      Lebani, Gianluca  and
      Lenci, Alessandro",
    editor = "Calzolari, Nicoletta  and
      Huang, Chu-Ren  and
      Kim, Hansaem  and
      Pustejovsky, James  and
      Wanner, Leo  and
      Choi, Key-Sun  and
      Ryu, Pum-Mo  and
      Chen, Hsin-Hsi  and
      Donatelli, Lucia  and
      Ji, Heng  and
      Kurohashi, Sadao  and
      Paggio, Patrizia  and
      Xue, Nianwen  and
      Kim, Seokhwan  and
      Hahm, Younggyun  and
      He, Zhong  and
      Lee, Tony Kyungil  and
      Santus, Enrico  and
      Bond, Francis  and
      Na, Seung-Hoon",
    booktitle = "Proceedings of the 29th International Conference on Computational Linguistics",
    month = oct,
    year = "2022",
    address = "Gyeongju, Republic of Korea",
    publisher = "International Committee on Computational Linguistics",
    url = "https://aclanthology.org/2022.coling-1.360",
    pages = "4101--4112",
}

@article{merlo2023blackbirdlanguagematricesblm,
      title={{Blackbird language matrices (BLM), a new task for rule-like generalization in neural networks: Motivations and Formal Specifications}}, 
      author={Paola Merlo},
      year={2023},
      volume={arXiv: 2306.11444},
      journal={arXiv},
      primaryClass={cs.CL},
      url={https://arxiv.org/abs/2306.11444}, 
}

@article{schutze2013judgment,
  title={{Judgment data}},
  author={Sch{\"u}tze, Carson T and Sprouse, Jon and Podesva, Robert J and Sharma, Devyani},
  journal={Research methods in linguistics},
  pages={27--50},
  year={2013}
}

@article{raven1938,
  title = 	 {{Standardization of progressive matrices}},
author = 	 {Raven, John C. },
  
  journal = 	 {British Journal of Medical Psychology},
  year = 	 {1938},
  volume = 	 {19},
  pages = 	 {137--150}
}

@article{wilson2023abstract,
    title = "How Abstract Is Linguistic Generalization in Large Language Models? Experiments with Argument Structure",
    author = "Wilson, Michael  and
      Petty, Jackson  and
      Frank, Robert",
    journal = "Transactions of the Association for Computational Linguistics",
    volume = "11",
    year = "2023",
    address = "Cambridge, MA",
    publisher = "MIT Press",
    url = "https://aclanthology.org/2023.tacl-1.78/",
    doi = "10.1162/tacl_a_00608",
    pages = "1377--1395"
}

@inproceedings{kann-etal-2019-verb,
    title = {{Verb Argument Structure Alternations in Word and Sentence Embeddings}},
    author = "Kann, Katharina  and
      Warstadt, Alex  and
      Williams, Adina  and
      Bowman, Samuel R.",
    booktitle = "Proceedings of the Society for Computation in Linguistics ({SC}i{L}) 2019",
    year = "2019",
    url = "https://aclanthology.org/W19-0129",
    doi = "10.7275/q5js-4y86",
    pages = "287--297",
}

@book{Levin93,
  address = {Chicago and London},
  author = {Levin, Beth},
  publisher = {University of Chicago Press},
  title = {{English Verb Classes and Alternations A Preliminary Investigation}},
  year = 1993
}

@inproceedings{an-etal-2023-blm,
    title = {{BLM-AgrF: A New French Benchmark to Investigate Generalization of Agreement in Neural Networks}},
    author = "An, Aixiu  and
      Jiang, Chunyang  and
      A. Rodriguez, Maria  and
      Nastase, Vivi  and
      Merlo, Paola",
    booktitle = "Proceedings of the 17th Conference of the European Chapter of the Association for Computational Linguistics",
    month = may,
    year = "2023",
    address = "Dubrovnik, Croatia",
    url = "https://aclanthology.org/2023.eacl-main.99",
    pages = "1363--1374",
}

@inproceedings{merlo2023-findings,
    title = "Blackbird language matrices ({BLM}), a new task for rule-like generalization in neural networks: Can Large Language Models pass the test?",
    author = "Merlo, Paola",
    editor = "Bouamor, Houda  and
      Pino, Juan  and
      Bali, Kalika",
    booktitle = "Findings of the Association for Computational Linguistics: EMNLP 2023",
    month = dec,
    year = "2023",
    address = "Singapore",
    publisher = "Association for Computational Linguistics",
    url = "https://aclanthology.org/2023.findings-emnlp.546/",
    doi = "10.18653/v1/2023.findings-emnlp.546",
    pages = "8119--8152"
}

@inproceedings{samo-etal-2023-blm,
    title = {{BLM-s/lE: A structured dataset of English spray-load verb alternations for testing generalization in LLMs}},
    author = "Samo, Giuseppe  and
      Nastase, Vivi  and
      Jiang, Chunyang  and
      Merlo, Paola",
    editor = "Bouamor, Houda  and
      Pino, Juan  and
      Bali, Kalika",
    booktitle = "Findings of the Association for Computational Linguistics: EMNLP 2023",
    month = dec,
    year = "2023",
    address = "Singapore",
    publisher = "Association for Computational Linguistics",
    url = "https://aclanthology.org/2023.findings-emnlp.821",
    doi = "10.18653/v1/2023.findings-emnlp.821",
    pages = "12276--12287",
}

@article{opitz-etal2025,
    author = {Opitz, Juri and Wein, Shira and Schneider, Nathan},
    title = {Natural Language Processing RELIES on Linguistics},
    journal = {Computational Linguistics},
    volume = {51},
    number = {3},
    pages = {1009-1032},
    year = {2025},
    month = {09},
    issn = {0891-2017},
    doi = {10.1162/coli_a_00560},
    url = {https://doi.org/10.1162/coli_a_00560},
    eprint = {https://direct.mit.edu/coli/article-pdf/51/3/1009/2512115/coli_a_00560.pdf},
}

@article{bresnan-2016-lifetime,
    title = {{Lifetime Achievement Award: Linguistics: The Garden and the Bush}},
    author = "Bresnan, Joan",
    journal = "Computational Linguistics",
    volume = "42",
    number = "4",
    month = dec,
    year = "2016",
    address = "Cambridge, MA",
    publisher = "MIT Press",
    url = "https://aclanthology.org/J16-4001/",
    doi = "10.1162/COLI_a_00260",
    pages = "599--617"
}

@article{schepens-etal-2025-llmcorpora,
    author = {Schepens, Job and Woloszyn, Hanna and Marx, Nicole and Gagl, Benjamin},
    title = {{Can Large Language Models Generate Useful Linguistic Corpora?: A Case Study of the Word Frequency Effect in Young German Readers}},
    journal = {Open Mind},
    volume = {9},
    pages = {1597-1656},
    year = {2025},
    month = {10},
    issn = {2470-2986},
    doi = {10.1162/OPMI.a.30},
    url = {https://doi.org/10.1162/OPMI.a.30},
    eprint = {https://direct.mit.edu/opmi/article-pdf/doi/10.1162/OPMI.a.30/2560068/opmi.a.30.pdf},
}

@article{Civico2025,
  author    = {Marco Civico},
  title     = {{Linguistic statistical universals: comparing computer- and human-generated texts}},
  journal   = {International Journal of Digital Humanities},
  year      = {2025},
  volume    = {7},
  number    = {1},
  pages     = {1--37},
  doi       = {10.1007/s42803-025-00096-7},
  url       = {https://doi.org/10.1007/s42803-025-00096-7},
  issn      = {2524-7840}
}

@inproceedings{samo2026modellingmorphologyverbalparadigms,
    title = "Modelling the Morphology of Verbal Paradigms: A Case Study in the Tokenization of {T}urkish and {H}ebrew",
    author = "Samo, Giuseppe  and
      Merlo, Paola",
    editor = {Oflazer, Kemal  and
      K{\"o}ksal, Abdullatif  and
      Varol, Onur},
    booktitle = "Proceedings of the Second Workshop Natural Language Processing for {T}urkic Languages ({SIGTURK} 2026)",
    month = mar,
    year = "2026",
    address = "Rabat, Morocco",
    publisher = "Association for Computational Linguistics",
    url = "https://aclanthology.org/2026.sigturk-1.8/",
    doi = "10.18653/v1/2026.sigturk-1.8",
    pages = "82--94",
    ISBN = "979-8-89176-370-8"
}

@article{guo2025deepseekr1,
  title={{Deepseek-r1: Incentivizing reasoning capability in llms via reinforcement learning}},
  author={Guo, Daya and Yang, Dejian and Zhang, Haowei and Song, Junxiao and Zhang, Ruoyu and Xu, Runxin and Zhu, Qihao and others},
  journal={arXiv preprint},
  volume={arXiv:2501.12948},
  year={2025},
  url={https://arxiv.org/abs/2501.12948}
}

@inproceedings{sasano-etal-2013-automatic,
    title = "Automatic Knowledge Acquisition for Case Alternation between the Passive and Active Voices in {J}apanese",
    author = "Sasano, Ryohei  and
      Kawahara, Daisuke  and
      Kurohashi, Sadao  and
      Okumura, Manabu",
    editor = "Yarowsky, David  and
      Baldwin, Timothy  and
      Korhonen, Anna  and
      Livescu, Karen  and
      Bethard, Steven",
    booktitle = "Proceedings of the 2013 Conference on Empirical Methods in Natural Language Processing",
    month = oct,
    year = "2013",
    address = "Seattle, Washington, USA",
    publisher = "Association for Computational Linguistics",
    url = "https://aclanthology.org/D13-1121/",
    pages = "1213--1223"
}

@article{hopper1980transitivity,
  title={{Transitivity in Grammar and Discourse}},
  author={Hopper, Paul J and Thompson, Sandra A},
  journal={Language},
  pages={251--299},
  year={1980},
  publisher={JSTOR}
}

@article{fillmore2003framenet,
  author  = {Fillmore, Charles J. and Johnson, Christopher R. and Petruck, Miriam R. L.},
  title   = {{Background to FrameNet}},
  journal = {International Journal of Lexicography},
  volume  = {16},
  number  = {3},
  pages   = {235--250},
  year    = {2003}
}

@inproceedings{guillaume-2021-graph,
    title = {{Graph Matching and Graph Rewriting: GREW tools for Corpus Exploration, Maintenance and Conversion}},
    author = "Guillaume, Bruno",
    editor = "Gkatzia, Dimitra  and
      Seddah, Djam{\'e}",
    booktitle = "Proceedings of the 16th Conference of the European Chapter of the Association for Computational Linguistics: System Demonstrations",
    month = apr,
    year = "2021",
    address = "Online",
    publisher = "Association for Computational Linguistics",
    url = "https://aclanthology.org/2021.eacl-demos.21/",
    doi = "10.18653/v1/2021.eacl-demos.21",
    pages = "168--175"
}

@inproceedings{nivre2015towards,
  title={{Towards a Universal Grammar for Natural Language Processing}},
  author={Nivre, Joakim},
  booktitle={International conference on intelligent text processing and computational linguistics},
  pages={3--16},
  year={2015},
  organization={Springer}
}

@article{de2021universal,
  title={Universal dependencies},
  author={De Marneffe, Marie-Catherine and Manning, Christopher D and Nivre, Joakim and Zeman, Daniel},
  journal={Computational linguistics},
  volume={47},
  number={2},
  pages={255--308},
  year={2021},
  publisher={MIT Press One Rogers Street, Cambridge, MA 02142-1209, USA journals-info~…}
}

@article{griffiths2024bayes,
  title={{Bayes in the Age of Intelligent Machines}},
  author={Griffiths, Thomas L and Zhu, Jian-Qiao and Grant, Erin and Thomas McCoy, R},
  journal={Current Directions in Psychological Science},
  volume={33},
  number={5},
  pages={283--291},
  year={2024},
  publisher={Sage Publications Sage CA: Los Angeles, CA}
}

@inproceedings{futrell-etal-2019-neural,
    title = {{Neural Language Models as Psycholinguistic Subjects: Representations of Syntactic State}},
    author = "Futrell, Richard  and
      Wilcox, Ethan  and
      Morita, Takashi  and
      Qian, Peng  and
      Ballesteros, Miguel  and
      Levy, Roger",
    editor = "Burstein, Jill  and
      Doran, Christy  and
      Solorio, Thamar",
    booktitle = "Proceedings of the 2019 Conference of the North {A}merican Chapter of the Association for Computational Linguistics: Human Language Technologies, Volume 1 (Long and Short Papers)",
    month = jun,
    year = "2019",
    address = "Minneapolis, Minnesota",
    publisher = "Association for Computational Linguistics",
    url = "https://aclanthology.org/N19-1004/",
    doi = "10.18653/v1/N19-1004",
    pages = "32--42"
}

@inproceedings{xiang-etal-2021-climp,
    title = "{CL}i{MP}: A Benchmark for {C}hinese Language Model Evaluation",
    author = "Xiang, Beilei  and
      Yang, Changbing  and
      Li, Yu  and
      Warstadt, Alex  and
      Kann, Katharina",
    editor = "Merlo, Paola  and
      Tiedemann, Jorg  and
      Tsarfaty, Reut",
    booktitle = "Proceedings of the 16th Conference of the European Chapter of the Association for Computational Linguistics: Main Volume",
    month = apr,
    year = "2021",
    address = "Online",
    publisher = "Association for Computational Linguistics",
    url = "https://aclanthology.org/2021.eacl-main.242/",
    doi = "10.18653/v1/2021.eacl-main.242",
    pages = "2784--2790"
}

@article{suijkerbuijk2025/COLI_a_00559,
    author = {Suijkerbuijk, Michelle and Prins, Zoë and Kloots, Marianne de Heer and Zuidema, Willem and Frank, Stefan L.},
    title = {{BLiMP-NL: A Corpus of Dutch Minimal Pairs and Acceptability Judgments for Language Model Evaluation}},
    journal = {Computational Linguistics},
    volume = {51},
    number = {4},
    pages = {1267-1301},
    year = {2025},
    month = {12},
    issn = {0891-2017},
    doi = {10.1162/COLI_a_00559},
    url = {https://doi.org/10.1162/COLI_a_00559},
    eprint = {https://direct.mit.edu/coli/article-pdf/51/4/1267/2512113/coli_a_00559.pdf},
}

@inproceedings{wendler-etal-2024-llamas,
    title = {{Do Llamas Work in {E}nglish? On the Latent Language of Multilingual Transformers}},
    author = "Wendler, Chris  and
      Veselovsky, Veniamin  and
      Monea, Giovanni  and
      West, Robert",
    editor = "Ku, Lun-Wei  and
      Martins, Andre  and
      Srikumar, Vivek",
    booktitle = "Proceedings of the 62nd Annual Meeting of the Association for Computational Linguistics (Volume 1: Long Papers)",
    month = aug,
    year = "2024",
    address = "Bangkok, Thailand",
    publisher = "Association for Computational Linguistics",
    url = "https://aclanthology.org/2024.acl-long.820/",
    doi = "10.18653/v1/2024.acl-long.820",
    pages = "15366--15394"
}

@inproceedings{nastase-etal-2025-multilingual,
    title = {{Multilingual vs. Monolingual Transformer Models in Encoding Linguistic Structure and Lexical Abstraction}},
    author = "Nastase, Vivi  and
      Samo, Giuseppe  and
      Jiang, Chunyang  and
      Merlo, Paola",
    editor = "Bosco, Cristina  and
      Jezek, Elisabetta  and
      Polignano, Marco  and
      Sanguinetti, Manuela",
    booktitle = "Proceedings of the Eleventh Italian Conference on Computational Linguistics (CLiC-it 2025)",
    month = sep,
    year = "2025",
    address = "Cagliari, Italy",
    publisher = "CEUR Workshop Proceedings",
    url = "https://aclanthology.org/2025.clicit-1.78/",
    pages = "826--836",
    ISBN = "979-12-243-0587-3"
}

@article{reinhart1981pragmatics,
  title={Pragmatics and linguistics: An analysis of sentence topics},
  author={Reinhart, Tanya},
  journal={Philosophica},
  volume={27},
  year={1981},
  publisher={Department of Philosophy and Moral Science at Ghent University.}
}

@phdthesis{volpato2010acquisition,
  title={{The Acquisition of Relative Clauses and phi-Features: Evidence from Hearing and Hearing-impaired Populations}},
  author={Volpato, Francesca},
  year={2010},
  school={Universit{\`a} Ca'Foscari Venezia},
}

@ARTICLE{brunatoetal2022/fpsyg.2022.707630,
    
AUTHOR={Brunato, Dominique  and Dell'Orletta, Felice  and Venturi, Giulia },
           
TITLE={{Linguistically-Based Comparison of Different Approaches to Building Corpora for Text Simplification: A Case Study on Italian}},
          
JOURNAL={Frontiers in Psychology},
          
VOLUME={Volume 13 - 2022},
  
YEAR={2022},
  
URL={https://www.frontiersin.org/journals/psychology/articles/10.3389/fpsyg.2022.707630},
  
DOI={10.3389/fpsyg.2022.707630},
  
ISSN={1664-1078}}

@book{Burzio1986,
  author    = {Burzio, Luigi},
  title     = {Italian Syntax: A Government and Binding Approach},
  year      = {1986},
  publisher = {Reidel},
  address   = {Dordrecht}
}

@article{SamoMerlo2021,
  author    = {Samo, Giuseppe and Merlo, Paola},
  year      = {2021},
  title     = {Intervention effects in clefts: a study in quantitative computational syntax},
  journal   = {Glossa: a Journal of General Linguistics},
  volume    = {6},
  number    = {1},
  pages     = {145},
  doi       = {10.16995/glossa.5742},
  url       = {https://doi.org/10.16995/glossa.5742}
}

@inproceedings{jiang-etal-2024-blm,
    title = {{BLM-It - Blackbird Language Matrices for Italian: A CALAMITA Challenge}},
    author = "Jiang, Chunyang  and
      Samo, Giuseppe  and
      Nastase, Vivi  and
      Merlo, Paola",
    editor = "Dell'Orletta, Felice  and
      Lenci, Alessandro  and
      Montemagni, Simonetta  and
      Sprugnoli, Rachele",
    booktitle = "Proceedings of the 10th Italian Conference on Computational Linguistics (CLiC-it 2024)",
    month = dec,
    year = "2024",
    address = "Pisa, Italy",
    publisher = "CEUR Workshop Proceedings",
    url = "https://aclanthology.org/2024.clicit-1.125/",
    pages = "1135--1143",
    ISBN = "979-12-210-7060-6"
}

@inproceedings{josifoski-etal-2023-exploiting,
    title = {{Exploiting Asymmetry for Synthetic Training Data Generation: SynthIE and the Case of Information Extraction}},
    author = "Josifoski, Martin  and
      Sakota, Marija  and
      Peyrard, Maxime  and
      West, Robert",
    editor = "Bouamor, Houda  and
      Pino, Juan  and
      Bali, Kalika",
    booktitle = "Proceedings of the 2023 Conference on Empirical Methods in Natural Language Processing",
    month = dec,
    year = "2023",
    address = "Singapore",
    publisher = "Association for Computational Linguistics",
    url = "https://aclanthology.org/2023.emnlp-main.96/",
    doi = "10.18653/v1/2023.emnlp-main.96",
    pages = "1555--1574"
}

@article{chen2024diversitysyntheticdataimpact,
      title={{On the Diversity of Synthetic Data and its Impact on Training Large Language Models}}, 
      author={Hao Chen and Abdul Waheed and Xiang Li and Yidong Wang and Jindong Wang and Bhiksha Raj and Marah I. Abdin},
      year={2024},
  volume={arXiv: 2410.15226},
      journal={arXiv preprint},
      primaryClass={cs.CL},
      url={https://arxiv.org/abs/2410.15226}, 
}

@inproceedings{lehmann-etal-1996-tsnlp,
    title = {{TSNLP - Test Suites for Natural Language Processing}},
    author = "Lehmann, Sabine  and
      Oepen, Stephan  and
      Regnier-Prost, Sylvie  and
      Netter, Klaus  and
      Lux, Veronika  and
      Klein, Judith  and
      Falkedal, Kirsten  and
      Fouvry, Frederik  and
      Estival, Dominique  and
      Dauphin, Eva  and
      Compagnion, Herve  and
      Baur, Judith  and
      Balkan, Lorna  and
      Arnold, Doug",
    booktitle = "{COLING} 1996 Volume 2: The 16th International Conference on Computational Linguistics",
    year = "1996",
    url = "https://aclanthology.org/C96-2120/"
}


\appendix
\onecolumn
\section*{Appendix}

\begin{table}[h]
\centering
\small
\begin{tabularx}{\textwidth}{l l l X X}
\hline
\toprule
 \textsc{$\#$Args} & \textsc{Voice} & \textsc{ST} & \textsc{Pattern} & \textsc{Without} \\ 
\midrule
\hline

2 & Act & Q &
V -[nsubj]-> Ag; V-[obj]-> Pat; Q [form="?"] &
Y [upos=VERB] \\

1& Act & Decl &
V -[nsubj]-> Ag; V-[obj]-> Pat &
Q [form="?"]; Y [upos="VERB"] \\

2 & Act & Q &
V -[nsubj]-> Ag; Q [form="?"] &
V-[obj]-> Pat; Y [upos="VERB"] \\

1 & Act & Decl &
V -[nsubj]-> Ag &
V-[obj]-> Pat; Q [form="?"]; Y [upos="VERB"] \\

2 & Pass & Q &
V -[nsubj:pass]-> Pat; V-[obl:agent]-> Ag; Q [form="?"] &
Y [upos="VERB"] \\

1 & Pass & Decl &
V -[nsubj:pass]-> Pat; V-[obl:agent]-> Ag &
Q [form="?"]; Y [upos="VERB"] \\

1 & Pass & Q &
V -[nsubj:pass]-> Pat; Q [form="?"] &
V-[obl:agent]-> Ag; Y [upos="VERB"] \\

1 & Pass & Decl &
V -[nsubj:pass]-> Pat &
V-[obl:agent]-> Ag; Q [form="?"]; Y [upos="VERB"] \\

\hline 

\end{tabularx}
\caption{Queries to retrieve natural instatiation example from UD treebanks with \textsc{Grew-Match} \citep{guillaume-2021-graph}. \textsc{$\#$Args} = number of arguments, \textsc{ST} = Sentence Type, Act = Active Voice, Pass = Passive Voice, Q = question, D = declarative. }
\label{table-queries-app}
\end{table}

\end{document}